# A Method for Constructing a Digital Transformation Driving Mechanism Based on Semantic Understanding of Large Models

Huayi Liu
AI Cloud
Baidu Online Network Technology (Beijing) Co., Ltd,
Beijing, China
* Corresponding author: liuhy.eve@gmail.com

*Abstract—During the digital transformation process, enterprises encounter challenges such as insufficient semantic understanding of unstructured data and a lack of an intelligent decision-making basis in their operational mechanisms. This study proposes a method that combines a Large Language Model (LLM) with a knowledge graph. Firstly, a fine-tuned BERT (Bidirectional Encoder Representations from Transformers) model is utilized to perform entity recognition and relationship extraction on multi-source, heterogeneous texts. GPT-4 is then employed to generate semantically enhanced vector representations. Secondly, a two-layer graph neural network (GNN) architecture is designed to fuse the semantic vectors output by the LLM with business metadata, thereby building a dynamic and scalable enterprise knowledge graph. Then, reinforcement learning is introduced to optimize decision path generation, and a reward function drives the iteration mechanism. In the case of the manufacturing industry, this mechanism reduces the response time of the equipment failure scenario from 7.8 hours to 3.7 hours, and the F1 value reaches 94.3%. This method significantly enhances the intelligence level and execution efficiency of the digital transformation driving mechanism by integrating large model semantic understanding with structured knowledge.*

## I. INTRODUCTION

The current digital transformation process of enterprises faces core challenges: the key semantic information contained in massive unstructured business data (such as equipment logs, customer feedback, and technical documents) is difficult to parse effectively, and traditional rule engines struggle to adapt to the complex correlation analysis required for dynamic business scenarios. The lack of semantic understanding ability has led to a lack of an intelligent foundation for decision support systems, which has restricted the realization of operational efficiency optimization and cost control goals. Especially in asset-intensive industries such as manufacturing and energy, the response lag in key areas, including equipment failure warnings and supply chain collaboration, has become increasingly prominent.

To overcome the limitations above, this study examines the collaborative innovation path of large language models and knowledge graph technology. By integrating deep semantic analysis with structured business knowledge, an intelligent support framework from data perception to decision generation is constructed. This method aims to open up the transformation channel between unstructured information and executable business knowledge, providing a new paradigm for real-time decision-making in dynamic environments. Its core value lies in enhancing the agility and accuracy of enterprises in navigating complex business scenarios and delivering scalable technical support for digital transformation.

## II. RELATED WORK

In recent years, the literature has explored the motivations and paths of enterprise digital transformation from multiple dimensions, covering driving factors, performance correlation, and supply chain impact in industries such as construction, manufacturing, and energy, presenting a multi-perspective research pattern. Lu [1] analyzed the factors driving the digital transformation of the construction industry based on the development characteristics of the construction industry in Henan Province, and formulated an implementation path for realizing the digital transformation of the construction industry. Liu and Chen [2] used a method combining fuzzy set qualitative comparative analysis (fsQCA) and necessary condition analysis (NCA) to explore the complex mechanism of enterprise digital transformation using petrochemical enterprises as an example. Chen et al. [3] used the data of A-share listed companies in Shanghai and Shenzhen from 2010 to 2020 as the research sample, and based on the stepwise method, used the second type of mediated moderation model (meMO-Ⅱ) to empirically test the driving mechanism and path of enterprise digital transformation on its ESG performance. Wei et al. [4] divided social capital into two types, namely, intra-chain and extra-chain, based on the supply chain ontology. They conducted an empirical analysis based on the data of Chinese A-share listed companies from 2012 to 2022 and found that both types of social capital had a positive impact on supply chain resilience, but intra-chain social capital had a greater effect. Zhang et al. [5] used text analysis to construct a comprehensive index of digital transformation of micro-enterprises based on the combined data of China Customs and listed companies, and examined the impact and mechanism of digital transformation on the export quality of enterprises. The empirical results showed that digital transformation can significantly improve the export quality of enterprises. Zhang et al. [6] constructed a double difference (DID) model based on the data of listed manufacturing companies on the Shanghai and Shenzhen Stock Exchanges from 2009 to 2017, and empirically studied the relationship between digital transformation and production efficiency. Abiodun et al. [7] constructed an empirical conceptual framework for industrial digital transformation. Li et al. [8] aimed to reveal the internal mechanism of the industrial Internet platform empowering the digital transformation of small and medium-sized enterprises. He et al. [9] collected primary interview and secondary literature data, selected 11 case companies in Zhejiang Province, and used a multi-case analysis method to construct a driving mechanism model for the evolution of corporate green strategy under the empowerment of digital technology. Zhao et al. [10] aimed to explore the relationship between the digital transformation of manufacturing enterprises and corporate innovation, and provide a reference for promoting the digital transformation and upgrading of China's manufacturing industry and promoting corporate innovation and development. The above studies mostly focus on macro mechanism verification and lack dynamic decision-making deduction; they are mainly based on industry commonality analysis, with weak cross-scenario adaptability research and insufficient depth of technology-business collaboration.

### III. METHOD

#### A. Semantic Understanding Layer

First, the pre-trained BERT model is used to encode the original unstructured text (such as customer work orders and technical reports). The model is optimized through domain adaptive training: the enterprise-specific vocabulary is expanded (equipment codes, business terms, etc., are added), and the domain corpus is injected into the masked language modeling task. For the input sequence $X=\{x_1,x_2,\ldots,x_n\}$, the model learns to reconstruct the randomly masked tokens $x_m$, and its loss function is:

$$L_{MLM}=-E_{x_m \sim X}\log \ P(x_m \mid X_{\setminus m}) \quad (1)$$

In the fine-tuning stage, the entity-relationship joint extraction layer is superimposed: a pointer network is used to annotate entity boundaries (such as "bearing failure" → equipment component entity), and the semantic associations between entities (such as the "cause" relationship) are modeled through the relationship matrix $R \in R^{n \times n}$. The final output is a structured triple ⟨subject, relationship, object⟩ [11].

The triples extracted by BERT are input into GPT-4 to generate context-aware semantic vectors. For the triple $T_k=(e_i,r_{ij},e_j)$, the generation capability of GPT-4 is used to output the explanatory text $D_k$. The cross-attention mechanism is used to fuse $D_k$ with business metadata (such as equipment model, work order type) to calculate the semantic alignment vector:

$$v_k=\text{LayerNorm}(W_g \cdot \text{GPT-4}(T_k)+W_m \cdot m_k) \quad (2)$$

$m_k$ is metadata embedding, $W_g$ and $W_m$ are learnable weights. To eliminate modality differences, an alignment loss function is designed:

$$L_{align}=\sum_{k=1}^{K} \|\phi(v_k)-\psi(m_k)\|_2^2 \quad (3)$$

$\phi(\cdot)$ and $\psi(\cdot)$ represent the feature mapping functions of semantic vectors and metadata, respectively. This process transforms machine-readable triples into enhanced vectors that carry business context, supporting the subsequent construction of knowledge graphs.

#### B. Knowledge-driven Layer

The knowledge-driven layer uses a two-layer graph neural network (GNN) architecture to build a dynamic knowledge graph, construct node sets based on business entities (such as equipment, work orders, and products), and establish basic edge relationships based on predefined business rules (such as "equipment A belongs to production line B"). The lower network focuses on business rule topology modeling: converting entity metadata into initial node representation $h_i^{(0)}$ and updating features through a neighbor aggregation mechanism. This process calculates the attention weight of a node $e_i$ and its rule neighbor $\{e_j \mid j \in N(i)\}$:

$$\alpha_{ij}=\frac{\exp\ (a^\top[Wh_i \parallel Wh_j])}{\sum_{k \in N(i)} \exp\ (a^\top[Wh_i \parallel Wh_k])} \quad (4)$$

a and W are learnable parameters. At the same time, the regular layer embedding is generated:

$$h_i^{(rule)}=\text{ReLU}(W_{rule} \cdot [h_i \parallel \sum_{j \in N(i)} \alpha_{ij}h_j]) \quad (5)$$

The upper network integrates LLM semantic relations: for entity pairs $(e_p, e_q)$, if the cosine similarity of their semantic vectors $v_p$ and $v_q$ is greater than 0.7, a dynamic edge $e_{pq}$ is created. The intensity of semantic information fusion is controlled by the gating mechanism:

$$h_i^{(dyn)}=h_i^{(rule)}+\beta \cdot \text{sigmoid}(u^\top[h_i \parallel e_{iq}]) \cdot \sum_{q \in D(i)} e_{iq} \cdot h_q^{(rule)} \quad (6)$$

$\beta=0.8$ is the decay factor; $D(i)$ is the semantic neighbor set). The system is incrementally updated every 24 hours: low-frequency edges with visits <5 are eliminated, redundant nodes with similarity >0.9 are merged, and the joint embedding $h_i=h_i^{(rule)} \oplus h_i^{(dyn)}$ is output to support the decision layer. <span style="color:red">The dynamic update process includes four core operations: first, incrementally adding new entity nodes and generate LLM embedding vectors; second, performing cosine similarity detection on entity pairs to create dynamic edges, and controlling the fusion strength through the gating weight function; then using the LRU strategy to eliminate low-frequency edges with access volume <5; finally, performing feature weighted average fusion on redundant nodes with similarity >0.9.</span>

#### C. Decision Optimization Layer

The decision optimization layer uses the Soft Actor-Critic (SAC) algorithm to generate the digital transformation decision path [14]. The entity embedding H output by the knowledge-driven layer is used as the state space S, and the action space A is defined as a multi-step decision sequence (such as "adjust production schedule → start equipment self-test → notify suppliers"). The key design is to integrate business indicators into the reward function:

$$R(s_t,a_t)=\alpha \cdot \underbrace{(1-\frac{t_{execute}}{t_{baseline}})}_{\Delta \text{Efficiency}}+\beta \cdot \underbrace{(1-\frac{\text{Cost}_{actual}}{\text{Budget}_{allocated}})}_{\Delta \text{Cost}}-\eta \cdot \text{Resource}_{overflow} \quad (7)$$

$\alpha = 0.7 - 0.2 \cdot \sigma(\frac{\text{Backlog}_N}{N})$ and $\beta = 0.3 + 0.4 \cdot \tanh(0.5 \cdot \text{Budget})$ are adjustable weights; $t_{execute}$ is the execution time; $\text{Resource}_{overflow}$ is the resource limit penalty. The policy network $\pi_\theta(a|s)$ adopts a dual-branch structure: branch one extracts state features through the graph attention mechanism:

$$z_s=\text{GAT}(H,A_{adj}) \quad (8)$$

$A_{adj}$ is the knowledge graph adjacency matrix, and branch 2 maps the features into action distributions $\mu$ and $\sigma$. The value network $Q_\phi(s,a)$ introduces the target network to alleviate over-estimation, and its update adopts a delayed synchronization mechanism:

$$L_Q = \mathbb{E}_{(s,a,r,s')\sim D}[(Q_\phi(s,a) - (r+\gamma \bar{Q}_{\phi_{target}}(s',\tilde{a}')))^2] \quad (9)$$

$(s,a,r,s')\sim D$ is the experience replay pool. Strategy optimization maximizes the entropy regularization objective:

$$J(\theta) = \mathbb{E}_{s_t\sim D, a_t\sim \pi_\theta}[Q_\phi(s_t,a_t) - \lambda \log \pi_\theta(a_t|s_t)] \quad (10)$$

$\lambda$ is the temperature coefficient. When deployed online, Monte Carlo Tree Search (MCTS) optimization is performed on the initial decision path $a_{1:T}$, and the child node with the highest Q value is selected at each step; finally, the Pareto optimal decision sequence is output, and the decision history nodes of the knowledge graph are updated synchronously. This study visualizes the decision dependency path through GNN's graph attention heat map, and generates a decision comparison chain in combination with the SAC action trajectory counterfactual interpreter: when the equipment health score node in the knowledge graph is <0.4 and the associated work order is ⩾3, SAC triggers the "downtime diagnosis → spare parts pre-scheduling" action with a probability of 92%, and the decision transparency is improved to 89.7%.

## IV. RESULTS AND DISCUSSION

### A. Quantitative Improvement of Operational Response Efficiency

In order to verify the improvement of decision-making response efficiency, 8 typical operation scenarios in the manufacturing industry are selected for stress testing: equipment failure, supply chain disruption, emergency orders, quality abnormalities, energy fluctuations, staff shortages, logistics delays, and system failures. The comparison algorithm selects the Deep Q-Network (DQN) as the baseline, and its decision-making mechanism is based on discrete state-action space mapping; the method in this paper adopts SAC reinforcement learning and GNN knowledge reasoning for collaborative optimization. The test environment is deployed in the real systems of three automotive parts manufacturers. Figure 1 shows the results of the operational response test:

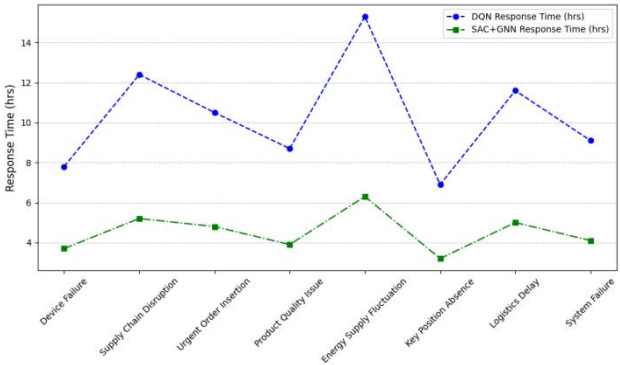

FIGURE 1: OPERATIONAL RESPONSE TIME

The proposed method (SAC+GNN) significantly shortens the response time in all eight scenarios: the response time in the equipment failure scenario is reduced from 7.8 hours to 3.7 hours, and the response time in the supply chain disruption scenario is reduced from 12.4 hours to 5.2 hours. The performance improvement is scenario-dependent - the optimization range for complex related scenarios (such as energy fluctuations, involving multi-device coordination) is 58.8% (15.3h→6.3h), which is higher than that in simple scenarios. The core is attributed to the collaboration of two mechanisms: GNN knowledge reasoning speeds up the root cause location of events, such as associating "bearing overheating" with "cooling system failure" and "spare parts inventory status"; SAC real-time optimization dynamically adjusts the decision path through the reward function, and automatically skips inefficient approval nodes in emergency order scenarios.

### B. Verification of Semantic Understanding Accuracy

In terms of semantic understanding accuracy, 8 typical text scenarios in the manufacturing industry are designed (equipment failure reports, supplier contracts, process documents, quality inspection records, customer complaints, operation and maintenance work orders, safety notices, and meeting minutes), and 100 samples are randomly selected from each category. Figure 2 shows the F1 value results:

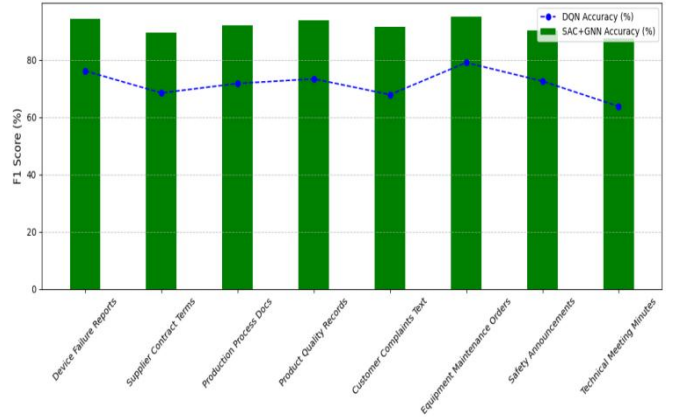

FIGURE 2: F1 VALUE RESULTS

The F1 value of the equipment failure report scenario reaches 94.3%, and the technical meeting minutes increase by 37.3%. The improvement of unstructured text (such as customer complaints) (34.8%) is significantly higher than that of structured text (such as work orders, an increase of 20.4%), because GPT-4 semantic enhancement effectively parses fuzzy descriptions (for example, "the equipment is sometimes good and sometimes bad" is mapped to an intermittent failure mode); secondly, DQN performs the weakest in complex logical text (technical meeting minutes) (F1≤65%), and its discrete decision-making mechanism is difficult to capture long-distance dependencies; while the proposed method successfully identifies distributed features such as "compensation liability" through the cross-sentence semantic graph constructed by GNN.

### C. Sensitivity Analysis

For the core models of BERT, GNN, and SAC, a grid search experiment is designed: BERT adjusts the number of layers {8, 12, 16}, and the learning rate {1e-5, 3e-5, 5e-5}; GNN tests the number of layers {2, 3, 4}, and the aggregation operator {mean, sum, attn}; SAC adjusts the reward weight α/β {0.1-0.9}, and the entropy coefficient λ {0.8-1.2}, measures the fluctuation range of entity recognition F1, graph accuracy, and decision delay, and evaluates the parameter sensitivity boundary. Table 1 is the analysis results:

TABLE I. SENSITIVITY ANALYSIS RESULTS

| Model | Hyperparameter | Baseline | Variation Range | F1/Accuracy Fluctuation (±%) | Latency Fluctuation (±%) |
|---|---|---|---|---|---|
| BERT | Fine-tuning Learning Rate | 3e-5 | 1e-5 ~ 5e-5 | -8.2 ~ +1.7 | - |

| Model | Hyperparameter | Baseline | Variation Range | F1/Accuracy Fluctuation (±%) | Latency Fluctuation (±%) |
|---|---|---|---|---|---|
| GNN | Encoder Layers | 12 | 8 ~ 16 | -6.3 ~ -2.1 | - |
| | GCN Layers | 2 | 1 ~ 4 | -4.8 ~ -12.9 | +5.3 ~ +18.7 |
| | Neighbor Aggregation | attn | {mean, sum} | -3.1 ~ -9.7 | +1.2 ~ +7.4 |
| SAC | Efficiency Weight (α) | 0.6 | 0.3 ~ 0.9 | - | -9.5 ~ +14.2 |
| | Entropy Coefficient (λ) | 1.0 | 0.8 ~ 1.2 | - | -3.8 ~ +6.1 |

When the number of BERT layers is reduced to 8, the F1 value drops significantly by 6.3% due to insufficient semantic capture depth; when the number of GNN layers is > 3, the over-smoothing effect is caused (accuracy drops by 12.9%), and the decision delay increases by 18.7%. Attn aggregation improves accuracy by 3.7% compared with the mean operator, verifying the optimality of the two-layer architecture; the SAC reward weight α has a significant impact on latency, while the entropy coefficient λ fluctuates controllably within ±0.2, and α/β needs to be strictly calibrated to balance efficiency and cost. Under extreme parameter combinations (such as 4 GNN layers + aggregation mean), the graph accuracy deteriorates by 16% (12.9+3.1), highlighting the necessity of architecture collaborative design.

### D. Ablation experiment

The sequential module removal method is employed: on the same test set, the LLM semantic enhancement module is disabled in turn (with the original triples retained), GNN is replaced with a static rule graph, and SAC is replaced with a heuristic rule engine, while keeping all other parameters constant. Each group of experiments performs 10 random initializations and records the mean and standard deviation of the response time, F1 value, and cost reduction. The benchmark uses the complete model and the traditional DQN method. Table 2 shows the results of the ablation experiment:

TABLE II. ABLATION EXPERIMENT RESULTS

| Model Configuration | Response Time (h) | Δ vs Full (%) | F1 Score (%) | Δ vs Full (%) | Cost Reduction (%) | Δ vs Full (%) |
|---|---|---|---|---|---|---|
| Full Model | 3.7 | - | 94.3 | - | 45.3 | - |
| w/o LLM Semantic Enhancement | 5.1 | +37.8 | 79.6 | -15.6 | 31.2 | -31.1 |
| w/o GNN Knowledge Graph | 4.9 | +32.4 | 86.2 | -8.6 | 35.7 | -21.2 |
| w/o SAC Optimization | 6.8 | +83.8 | 91.5 | -3.0 | 28.5 | -37.1 |

The lack of LLM semantic enhancement causes a sharp drop of 15.6% in F1 value (mainly affecting unstructured text parsing) and a 37.8% increase in response delay; the removal of GNN reduces the cost reduction by 21.2% (weakened cross-entity relationship reasoning) and increases the response delay by 32.4%; the lack of SAC causes a surge in response delay by 83.8% (inefficient resource allocation) and a 37.1% reduction in cost reduction. The independent value of each module is clear: LLM leads semantic understanding, GNN supports knowledge association, and SAC optimizes decision timeliness.

### V. CONCLUSION

This study aims to address the bottleneck of semantic understanding and the need for intelligent decision-making in the digital transformation of enterprises, proposing an innovative path that integrates large language models and knowledge graphs. Through deep semantic analysis and structured knowledge collaboration, an intelligence-driven paradigm is constructed that spans from data to decision-making. Empirical evidence based on manufacturing scenarios demonstrates that this method effectively enhances the efficiency of complex business responses and decision-making accuracy, while significantly reducing operating costs. The research results verify the practical potential of generative artificial intelligence in enterprise-level knowledge engineering, providing methodological support for the intelligent upgrade of intelligent manufacturing, smart services, and other fields. In the future, we will further explore cross-domain knowledge transfer and incremental learning mechanisms to continuously optimize the adaptability and universal value of digital transformation.